\documentclass[runningheads]{llncs}
\usepackage[T1]{fontenc}
\usepackage{graphicx}
\usepackage{booktabs}
\usepackage{amsmath}
\usepackage{amssymb}
\usepackage{url}
\usepackage{booktabs,tabularx}
\usepackage{mathtools} 
\usepackage{colortbl,xcolor}
\usepackage{multirow}
\usepackage{makecell}
\usepackage{array}
\usepackage{framed}
\usepackage[most]{tcolorbox}
\usepackage{enumitem}
\usepackage[misc]{ifsym}
\newcommand{\corr}{(\Letter)}
\usepackage{mwe}

\begin{document}

\title{Concept-SAE: A Controllable and Invertible Concept Interface for Sparse Autoencoders}

\titlerunning{Concept-SAE: Controllable Concept Interface for Sparse Autoencoders}

\author{Jianrong Ding\thanks{Equal contribution. \Letter \hspace{0.1em} Corresponding author.} \and
Muxi Chen$^\star$ \and
Chenchen Zhao \and
Qiang Xu\corr}

\authorrunning{J. Ding et al.}

\institute{The Chinese University of Hong Kong
\email{\{jrding25,mxchen21,cczhao,qxu\}@cse.cuhk.edu.hk}}

\maketitle              

\begin{abstract}
Standard Sparse Autoencoders (SAEs) excel at discovering a dictionary of a model’s learned features, providing a powerful lens for passive feature discovery. However, this passive nature makes it difficult to systematically evaluate or analyze concepts that users explicitly care about. We introduce Concept-SAE, a framework that augments SAEs with a structured and controllable interface for probing user-defined concepts. Concept-SAE decomposes an activation subspace into two orthogonal components: Concept Tokens, which are aligned to externally specified semantics through dual supervision on both concept existence and spatial localization, and Free Tokens, which operate like standard SAEs to capture all remaining information. This hybrid disentanglement strategy ensures that Concept Tokens are faithful, spatially grounded, and cleanly separated from the residual subspace while preserving the ability of SAEs for open-ended concept discovery. We conduct extensive experiments demonstrating that Concept-SAE yields high-fidelity, well-localized, and strongly disentangled concept representations, outperforming alternatives in interface quality. Finally, we validate the utility of this conceptual interface through three diagnostic evaluations: a detection test on classifying adversarial image samples, a controllability test focusing on controlled counterfactual editing and a stability test using adversarial perturbations. Together, these results show that Concept-SAE equips SAEs with a reliable mechanism for evaluating, probing, and diagnosing user-defined concepts. The code is available at \url{https://github.com/RafaDD/Concept-SAE}.
\keywords{Interpretability \and Sparse autoencoder}
\end{abstract}
\section{Introduction}

The ultimate goal of mechanistic interpretability~\cite{olah2018building,olah2020zoom} is to reverse-engineer neural networks, shifting from observing their behavior to understanding the internal algorithms they implement. A dominant line of work uses Sparse Autoencoders (SAEs)~\cite{huben2023sparse,ramaswamy2023overlooked,yeh2020completeness} to decompose a model’s activations into a dictionary of learned features. This approach has been widely applied in both vision~\cite{zhang2018visual,stevens2025sparse,lim2024sparse,olson2025probing} and language~\cite{shu2025survey,huben2023sparse}, providing a powerful lens for passive feature discovery.

Despite this promise, analyzing vision models with SAEs remains cumbersome and limited. Current practice requires manually inspecting SAE tokens, identifying images with strong activations, and visually inferring the underlying concepts~\cite{gao2024scaling,paulo2025sparse,marks2024enhancing,harle2025measuring}. In essence, such procedures passively check whether a model happens to encode human-specified concepts.

A natural next step is to move from passive observation to active test of concept usages. One straightforward strategy is to constrain a subset of SAE tokens to represent concept scores, akin to concept bottleneck models~\cite{koh2020concept}. However, binary presence signals provide limited information and fail to disentangle high-dimensional concepts, often bleeding semantics into unconstrained tokens. Another strategy embeds concepts in those tokens as vectors~\cite{espinosa2022concept,espinosa2023learning}, but without explicit supervision these embeddings tend to drift from their intended meanings, undermining interpretability and restricting the capacity for open-ended analysis of the original SAE. Overall, actively testing concept usage involves dedicated designs.

To bridge this gap, we introduce Concept-SAE, a framework that provides a structured and invertible conceptual interface for SAE. The goal of Concept-SAE is to equip SAEs with the ability to verify and manipulate whether—and how—a model represents a predefined set of human concepts, while preserving their capacity for passive feature discovery. Concept-SAE decomposes an activation subspace into two orthogonal components: (1) \textbf{Concept Tokens} aligned with human-defined semantics (e.g., beard, eyes). We apply dual supervision—on concept existence via scalar scores and on spatial localization via segmentation masks—to ensure that Concept Tokens are faithful, spatially grounded, and disentangled from the rest of the representation. (2) \textbf{Free Tokens} trained in the standard SAE manner to capture all remaining information. These tokens enable open-ended concept discovery and ensure a complete decomposition of the activation space.
The main contributions of this work are as follows:
\begin{itemize}
    \item We introduce Concept-SAE, a novel framework that equips SAE with the ability to align models' internal representations with external semantics, providing a high-fidelity, bidirectional interface for concept-based inspection.
    \item We provide extensive experimental validation showing our hybrid strategy produces concept representations that are remarkably faithful, spatially localized, and cleanly disentangled from the free-token residual space, demonstrating superiority over alternative methods in interface quality.
    \item We demonstrate our proposed concept scores accurately reflect model-internal certainty with an adversarial image sample detection test (Detection test), a controlled counterfactual editing experiment (Controllability Test), and an adversarial perturbations experiment (Stability Test).
\end{itemize}

\section{Related Works}
\subsection{Model Interpretability with Sparse Autoencoder}
Sparse Autoencoders (SAEs) have emerged as powerful tools for mechanistic interpretability, building on the sparse coding hypothesis~\cite{olshausen1997sparse} to address feature superposition. By training sparse decoders to reconstruct model activations, SAEs decompose polysemantic representations into more interpretable latent features~\cite{sharkey2022taking,huben2023sparse}. They have been applied across a range of architectures—including MLP blocks, attention heads~\cite{kissane2024interpreting}, and both visual~\cite{gorton2024missing} and textual models~\cite{kantamnenisparse,mudideefficient,minegishirethinking}—with subsequent work improving training stability and feature structure~\cite{rajamanoharan2024improving}. SAE-derived features now support tasks such as analyzing computational circuits~\cite{o2024sparse} and enabling targeted model interventions for safety~\cite{marks2023interpreting}.

While SAEs provide a strong observational tool, their use in vision models has largely remained passive. Analyses typically rely on visually inspecting latent tokens, selecting images with strong activations, and inferring the underlying semantics from activation maps~\cite{gao2024scaling,paulo2025sparse,marks2024enhancing,harle2025measuring}. This workflow effectively checks post hoc whether a model happens to use certain human-defined concepts, but offers limited ability to actively evaluate how specific concepts are represented or to directly probe their roles within the model.

In contrast, our approach aims to complement SAEs by providing an interface that enables active testing of user-defined concepts, while preserving the open-ended feature discovery capabilities that make SAEs valuable for mechanistic interpretability.

\subsection{Incorporating Predefined Concepts into SAE}
Our approach extends sparse autoencoders by explicitly incorporating predefined concepts into their latent space. Since SAEs are trained to reconstruct internal features, constraining certain tokens to represent human concepts parallels prior efforts that inject concepts into the prediction process. A representative example is the Concept Bottleneck Model (CBM)~\cite{koh2020concept,rao2024discover}, which supervises latent features to align with predefined concepts and forces predictions to pass through these human-understandable variables. Another line of work encodes concepts as latent embeddings~\cite{espinosa2022concept,espinosa2023learning}.  
However, these approaches are primarily designed for the output layer. In intermediate feature spaces, binary bottlenecks struggle to disentangle fine-grained concepts and often cause semantic overlap~\cite{espinosa2022concept}, while embeddings easily drift from their intended semantics without direct supervision~\cite{espinosa2022concept}. To address these challenges, we supervise the values of concept tokens, anchoring them to visual evidence that captures both existence and spatial localization. A staged training strategy further enforces this separation, preventing leakage into free tokens while preserving their exploratory capacity.

\begin{figure*}[!t]
    \centering
    \includegraphics[width=\linewidth]{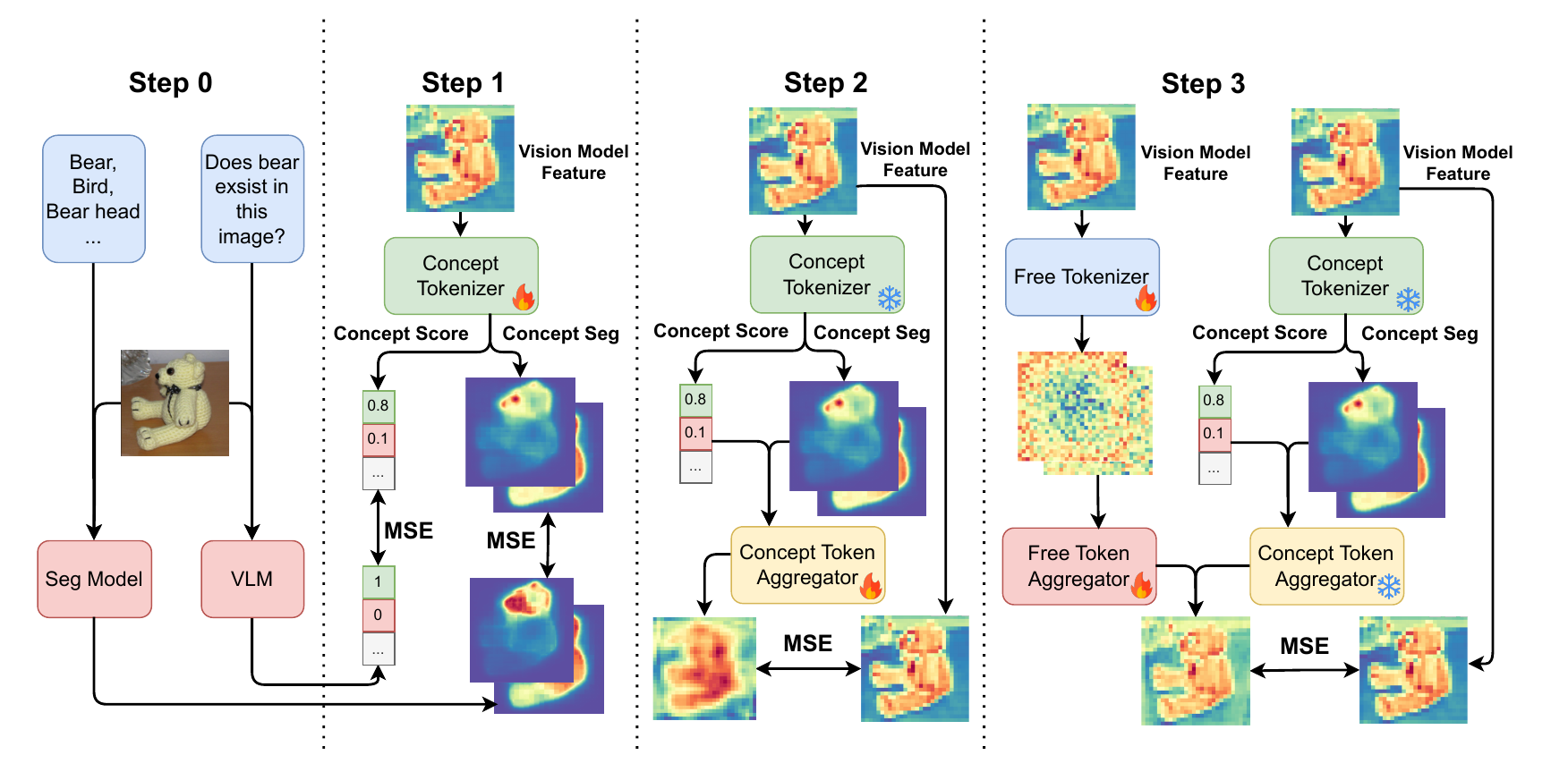}
    \caption{Overall training pipeline of our proposed method. The fire icon means the module is trainable, while the snow icon means the parameters of the module are frozen.}
    \label{fig:main_method}
\end{figure*}
\section{Methodology}

\textbf{Concept-SAE} endows SAE-based interpretability models with the capability of incorporating predefined concepts. The overall procedure begins with concept label generation, we first generate a series of concepts (e.g. bear, bird, bear head, etc.). Then we use a segmentation model to get the position mask of the concept in the image, and use a VLM to output binary labels indicating whether the concept exists in the figure. We then train a Concept Autoencoder, composed of a Concept Tokenizer and a Concept Aggregator: the tokenizer learns to extract concept segmentation and concept existence score from vision model features, while the aggregator reconstructs vision model features from the concept segmentation and concept existence score. Thereby, we form a fully invertible, concept-centric representation of the vision model’s internal state. To capture residual information beyond the predefined concept space, we further introduce a Free Autoencoder, which functions analogously to a conventional sparse autoencoder. The overall model training pipeline with four steps is shown in Fig.~\ref{fig:main_method}.

\subsection{Concept Label \& Segmentation Generation}
To incorporate predefined concepts without constraining the SAE’s ability to discover novel features, we adopt a minimal representation for each concept, consisting of its existence and spatial extent. This ensures that concept supervision anchors interpretation without injecting unnecessary information that could interfere with the SAE. We leverage two plug-and-play models to provide precise ground-truth annotations of these two signals: A vision-language model (VLM) determines whether each concept is present in the image, yielding a binary existence score. In parallel, a segmentation model produces an initial spatial mask. The two outputs are then fused: if the VLM judges a concept absent, its mask is suppressed to zero; otherwise, the original mask is preserved. This refinement yields clean and reliable annotations for training the concept-based modules.
\begin{figure*}[!t]
    \centering
    \includegraphics[width=\linewidth]{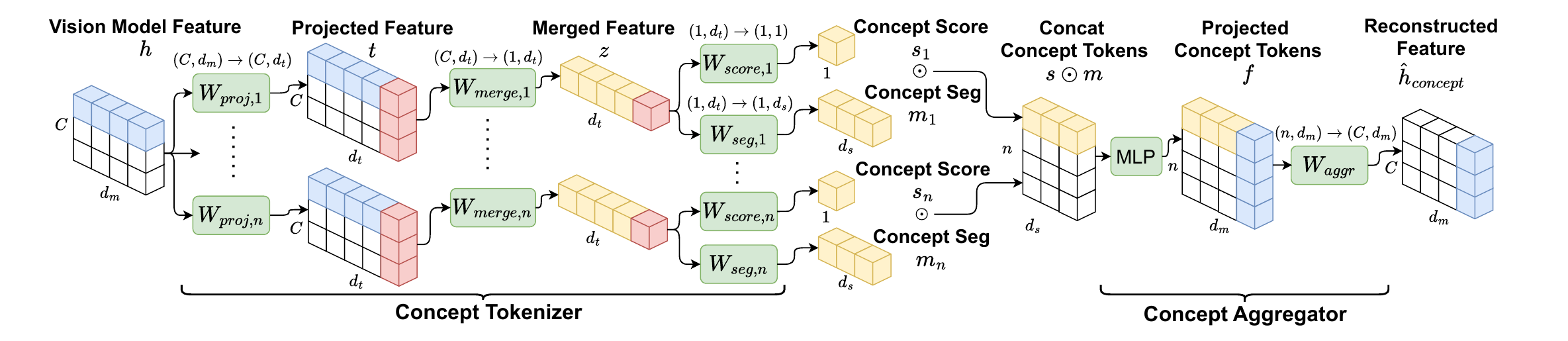}
    \caption{Computation process of concept tokenizer and concept aggregator.}
    \label{fig:computation}
\end{figure*}
\subsection{Concept Tokenizer}

\begin{figure*}[!t]
    \centering
    \includegraphics[width=\linewidth]{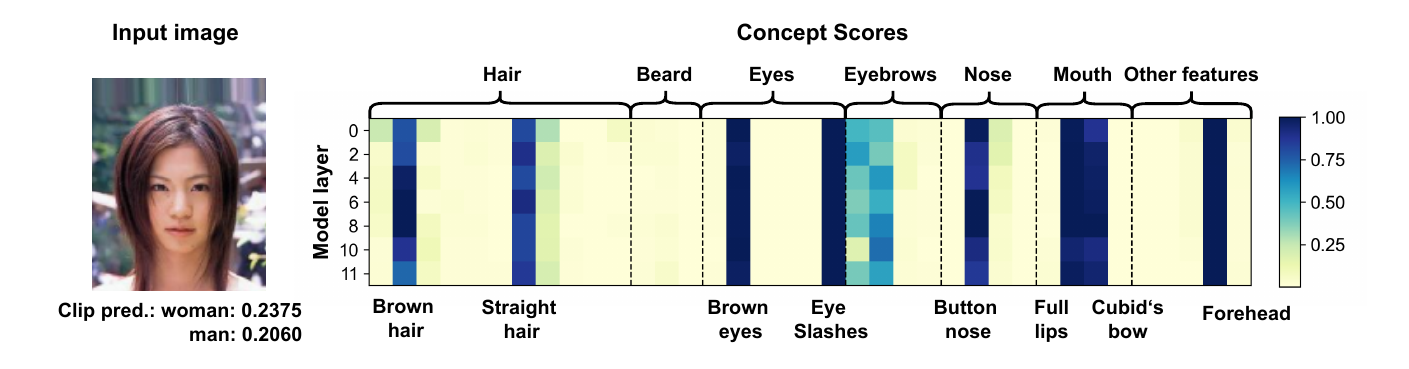}
    \caption{Concept score example for an image. The y-axis is the layer of the model, and x-axis is the concepts. The name of the concepts with a high concept score is shown at the bottom.}
    \label{fig:concept_score_show_1}
\end{figure*}

The Concept Tokenizer \(\mathcal{T}_{concept}\) serves as an encoder, it maps the hidden representations of the target model onto a constrained concept space. With the generated annotations, it is trained to predict two signals for each concept: a binary existence score and a spatial mask. As shown in the left part of Fig.~\ref{fig:computation}, for each concept, the tokenizer first projects the internal feature maps into a dedicated latent space through a learnable transformation \(W_{proj,i}\), producing concept-specific embeddings for \(n\) concepts. Each embedding is aggregated to form holistic concept representations with parameter \(W_{merge,i}\), which integrate evidence distributed over different channels or patches. Based on this unified representation, the tokenizer predicts two outputs: a binary concept score indicating whether the corresponding concept exists in the vision model feature and a spatial mask localizing the spatial position of concept within the vision model feature. We use two linear layers to compute the concept score and the concept segmentation separately (\(W_{score,i}\) and \(W_{seg,i}\)). The formulation of this process is shown in Eq.~\ref{eq:score_pred} and Eq.~\ref{eq:seg_pred}.
\begin{equation}
\label{eq:score_pred}
s_i = \text{Sigmoid}\left(z_i^{(d_t)} \cdot W_{score, i}^{(d_t)} + b_{score, i}\right), 1 \leq i \leq n
\end{equation}
\begin{equation}
\label{eq:seg_pred}
m_i^{(d_s)} = z_i^{(d_t)} \cdot W_{seg, i}^{(d_t \times d_s)} + b_{seg, i}^{(d_s)}, 1 \leq i \leq n
\end{equation}
where \(s\in \mathbb{R}^{n}\) is the predicted concept score for the internal feature, and \(m\in \mathbb{R}^{n\times d_s}\) is the predicted concept mask. The training loss is designed as the mean squared error (MSE) between the predicted concept score \(s\), predicted concept mask \(m\) and the true concept score \(\mathcal{S}\), concept masks \(\mathcal{M}\). We also apply an \(L_1\) penalty to \(W_{merge}\) to encourage sparsity, ensuring that each concept's representation is derived from a minimal set of channels or patches. The loss is formulated as Eq.~\ref{eq:tokenizer_loss}.
\begin{equation}
    \mathcal{L}_{tokenizer} = \lambda_1||\mathcal{S} - s||_2^2 + \lambda_2||\mathcal{M} - m||_2^2 + \lambda_3||W_{merge}||_1
\label{eq:tokenizer_loss}
\end{equation}
By doing so, \(\mathcal{T}_{concept}\) enforces a direct alignment between intermediate features and human-interpretable concepts. An example of the predicted concept scores for different layers of a model is shown in Fig.~\ref{fig:concept_score_show_1}.

\subsection{Concept Aggregator}

With the trained \(\mathcal{T}_{concept}\), we freeze its parameters and further train a Concept Aggregator \(\mathcal{A}_{concept}\), which is the decoder that reconstructs the original vision model feature \(h\) from the predicted concept representations. The right part of Fig.~\ref{fig:computation} shows its computation process in two steps. Firstly, it combines the information from the predicted concept score and segmentation by element-wise multiplication. The segmentation features of concepts absent from the image will be masked by low concept scores. Then, we utilize an MLP to fuse the predicted concept score \(s\) and segmentation \(m\) into a unified feature vector \(f \in \mathbb{R}^{n\times d_m}\). Finally, we combines these concept features with a fully connected layer to produce the reconstructed feature map \(\hat{h}_{concept}\).
The training loss is designed as the MSE between the predicted feature \(\hat{h}_{concept}\) and the true vision model feature \(h\). Meanwhile, we want to encourage that every concept contributes to the channels or patches where it is retrieved. Therefore, we add a KL divergence loss between \(W_{aggr}\) in \(\mathcal{A}_{concept}\) and \(W_{merge}\) in \(\mathcal{T}_{concept}\), so that concept-channel distributions of the two parameters are aligned. We also add the \(L_1\) loss to \(W_{aggr}\) to encourage sparsity: each concept only contributes to the reconstruction of a few channels or patches. The loss function of \(\mathcal{A}_{concept}\) is shown in Eq.~\ref{eq:aggregator_loss}, where \(\text{sm}(\cdot)\) is the softmax function.
\begin{equation}
    \mathcal{L}_{aggr} = \lambda_1||\hat{h}_{concept} - h||_2^2  + \lambda_2||W_{aggr}||_1 + \lambda_3\text{KL}(\text{sm}(W_{merge}) ||\text{sm}(W_{aggr}^\top) )
\label{eq:aggregator_loss}
\end{equation}

\subsection{Free Tokenizer \& Free Aggregator}

\begin{figure*}[!t]
    \centering
    \includegraphics[width=\linewidth]{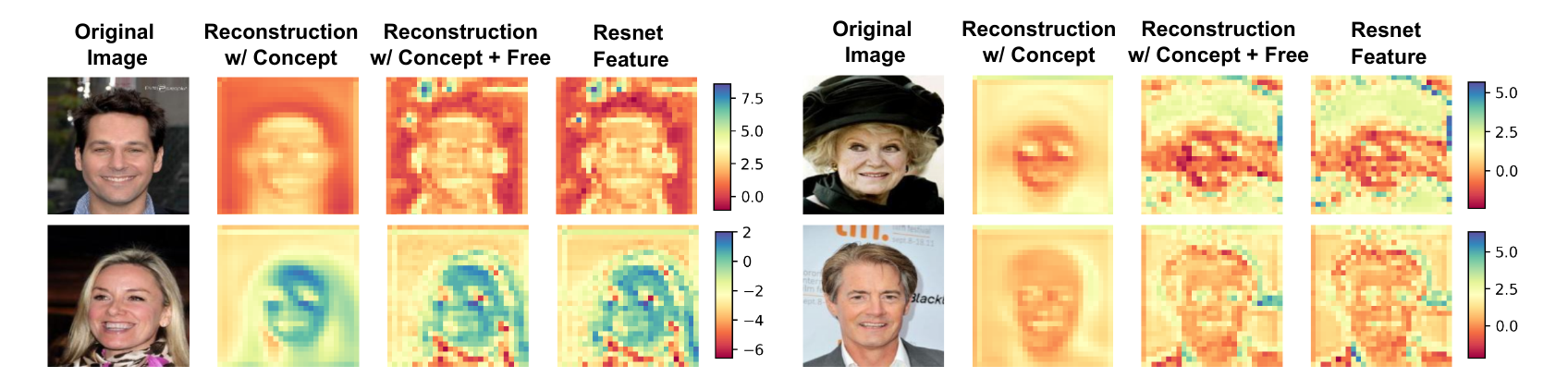}
    \caption{Reconstruction examples of our proposed method on ResNet features.}
    \label{fig:reconstruction_show_1}
\end{figure*}

We introduce a Free Tokenizer \(\mathcal{T}_{free}\) and a Free Aggregator \(\mathcal{A}_{free}\) to discover features not covered by the predefined concept space. These two modules share the same architecture as the concept tokenizer–aggregator pair, but differ in training strategies: unlike the concept modules, they are trained jointly under the objective of the SAE with no external supervision. 
The training loss for the free tokenizer and the free aggregator is designed to encode implicit concepts absent from the predefined concept pool and reconstruct the original features jointly with \(\mathcal{T}_{concept}\) and \(\mathcal{A}_{concept}\). The combination of the reconstructed features from the concept module and the free module is performed by a simple adding operation. Meanwhile, the \(L_1\) loss is added to the output of the free tokenizer to preserve sparsity. The detailed loss is formulated in Eq.~\ref{eq:free_loss}, with \(\mathcal{T}_{free}\) and \(\mathcal{A}_{free}\) as the free tokenizer and the free aggregator separately. A comparison of reconstruction performance between the joint modules and the concept modules alone is shown in Fig.~\ref{fig:reconstruction_show_1}.
\begin{equation}
\label{eq:free_loss}
\mathcal{L}_{free} = \lambda_1||\mathcal{A}_{free}(\mathcal{T}_{free}(h)) + \hat{h}_{concept} - h||_2^2 + \lambda_2||\mathcal{T}_{free}(h)||_1
\end{equation}

\section{Experiments}

To validate our proposed method, we conduct a series of experiments to investigate (1) whether the representations of concepts derived by Concept-SAE are faithful and disentangled, and (2) Fidelity of the representations of Concept-SAE. We verify the fidelity of the representation via downstream tasks, including model interpretation, counterfactual editing, and robustness analysis. Our evaluation is guided by the following research questions:

\begin{itemize}
    \item \textbf{RQ1: Concept Focus \& Faithfulness.} Comparing to standard SAEs and concept-embedding baselines, do concept tokens faithfully capture model-internal, human-interpretable concepts while remaining sparse and disentangled?
    \item \textbf{RQ2: Diagnostic Utility via Detection.} Do the extracted concept scores accurately reflect internal model certainty, and can the ambiguity in these representations be leveraged to explain prediction failures and effectively detect adversarial image samples?
    \item \textbf{RQ3: Diagnostic Utility via Controllability.} Is the fidelity of the Concept-SAE representation high enough to enable reliable, controllable counterfactual editing, allowing us to causally intervene on specific concept scores to correct model mispredictions?
    \item \textbf{RQ4: Diagnostic Utility via Stability Diagnosis.}  Is the semantic alignment of our concept tokens robust enough to serve as a semantic stability test, where concept distortions under adversarial attack provide meaningful diagnostics for localizing and mitigating model vulnerabilities?
\end{itemize}

\subsection{Experimental Setup}

\textbf{Datasets.} We evaluate our approach on two datasets: \textbf{(1) CelebA}~\cite{liu2015faceattributes}, we focus on the binary classification of the \textit{Gender} attribute as the target label; and \textbf{(2) ImageNet-1k}~\cite{deng2009imagenet}, which involves classification across 1000 object categories. 

\noindent\textbf{Models.} We consider two representative vision architectures: \textbf{(1) ResNet-18}~\cite{he2016deep}, which is trained on CelebA and ImageNet-1k separately and subsequently analyzed; and \textbf{(2) Vision Transformer (ViT-B/32)}~\cite{dosovitskiy2020image,radford2021learning}, which is pre-trained on the LAION-2B dataset, and we evaluate the model with Concept-SAE in a zero-shot setting on both CelebA and ImageNet-1k.

\subsection{Detailed Training Setup and Hyperparameters}
\label{sec:train_details}

We adopt a three-stage training strategy for our proposed framework: \textbf{(1)} training the Concept Tokenizer, \textbf{(2)} training the Concept Aggregator, and \textbf{(3)} training the Free Tokenizer and Free Aggregator. All experiments are conducted on an Ubuntu 22.04 server equipped with an AMD EPYC 7K62 CPU and an NVIDIA A100-64G GPU. Below we describe each training stage in detail.

\textbf{Concept Tokenizer Training.} We use Adam as optimizer with an initial learning rate of \(1\times10^{-3}\), scheduled by step learning rate scheduler with \(\gamma=0.1\) and step size 20, and train for 30 epochs. A batch size of 64 is applied. The loss function is shown below, combining existence score error, mask error, and an \(L_1\) sparsity penalty, weighted by coefficients \((\lambda_1, \lambda_2, \lambda_3)=(1,\hspace{0.25em} 1,\hspace{0.25em} 0.1)\).

\textbf{Concept Aggregator Training.} This stage uses Adam as optimizer with learning rate \(1\times10^{-3}\), scheduled by Step learning rate scheduler with \(\gamma=0.1\) and step size 30, with batch size 64. It is trained for 50 epochs. The loss function is shown below. Our proposed loss combines feature reconstruction error, KL divergence aligning \(W_{merge}\) and \(W_{aggr}\), and an \(L_1\) penalty on aggregator weights,  weighted by coefficients \((\lambda_1, \lambda_2, \lambda_3)=(1,\hspace{0.25em} 0.01,\hspace{0.25em} 1)\).

\textbf{Free Tokenizer and Free Aggregator Training.} Finally, we set the number of free tokens to be 36 for both the CelebA and the ImageNet dataset. We use Adam optimizer with a learning rate of \(1\times10^{-3}\). It is trained for 30 epochs with a batch size of 64. The loss function is shown below. The loss objective combines reconstruction error with an $L_1$ sparsity penalty, weighted by \((\lambda_1, \lambda_2)=(1,\hspace{0.25em} 1)\).

\begin{figure*}[!t]
    \centering
    \includegraphics[width=\linewidth]{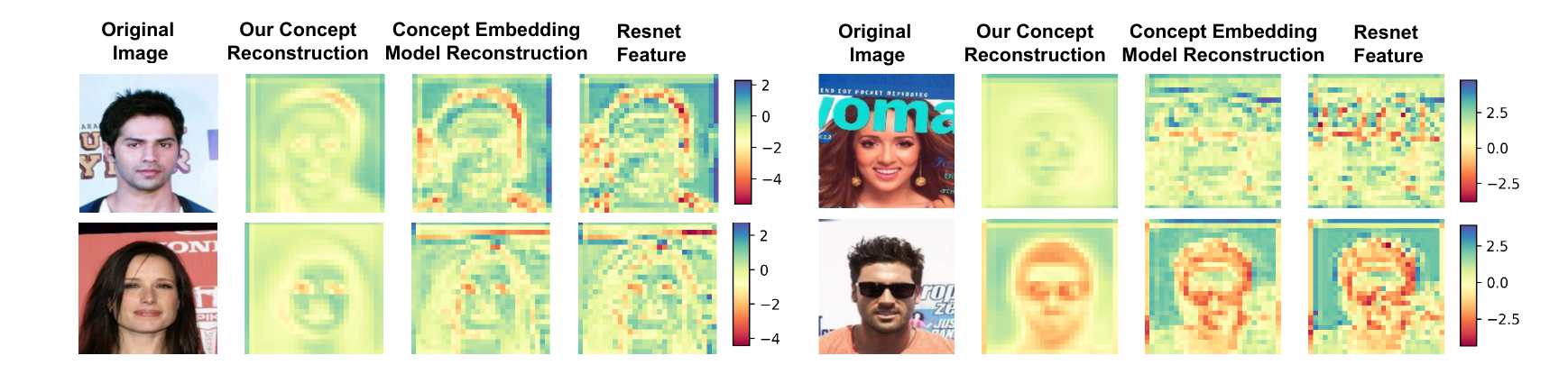}
    \caption{The difference between the vision model feature reconstructed by our concept module and concept embedding model.}
    \label{fig:reconstruction_contrast_1}
\end{figure*}

\begin{figure*}[!t]
    \centering
    \includegraphics[width=\linewidth]{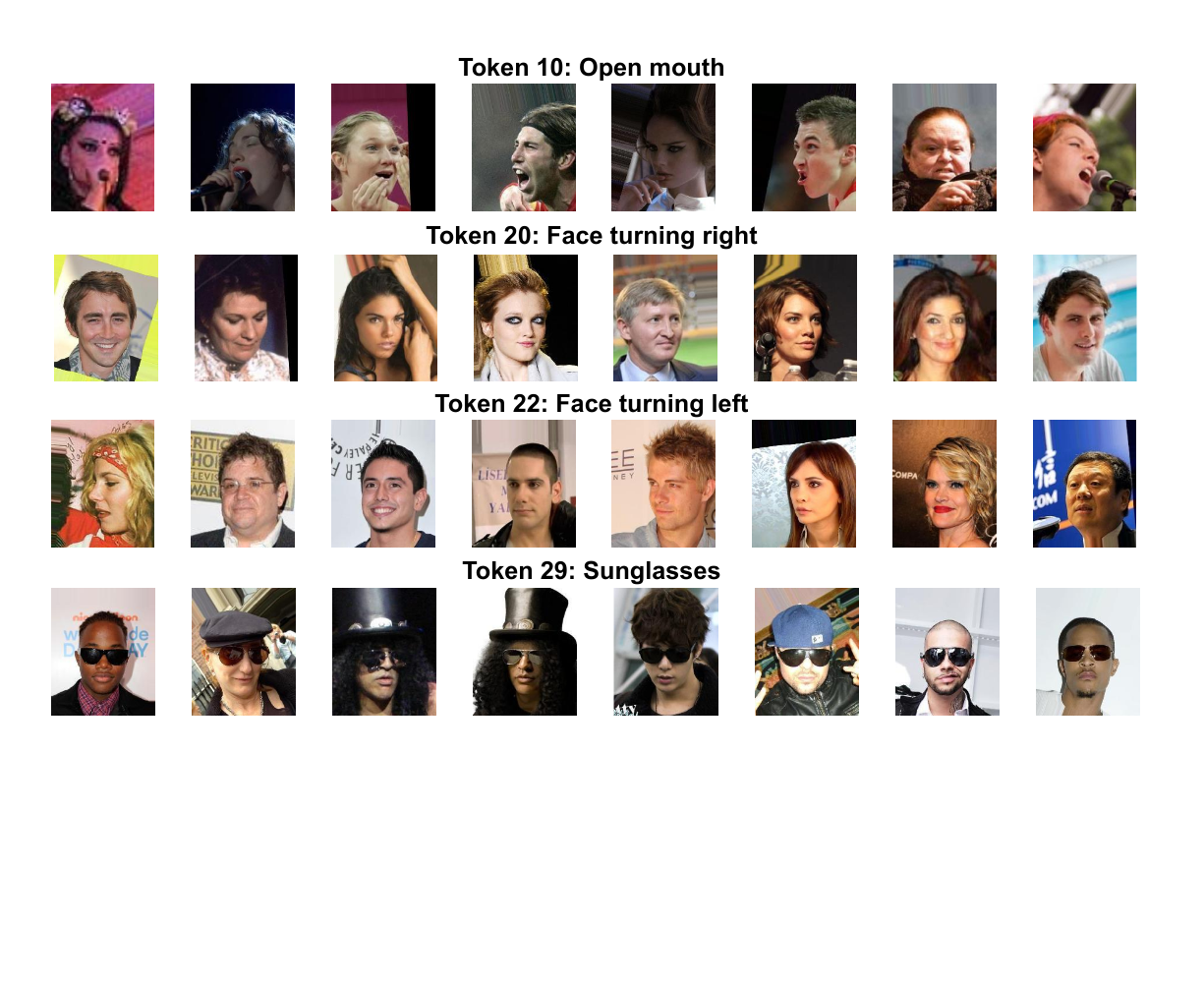}
    \caption{Four examples of free token visualizations, each with 8 most activated images from the CelebA dataset.}
    \label{fig:free_token_part}
\end{figure*}

\subsection{RQ1: Concept Focus \& Faithfulness}
We evaluate the purity of our learned concepts by comparing them with a concept-embedding baseline (CEM)~\cite{espinosa2022concept}, which represents each concept with two complementary embeddings supervised only by concept existence. As shown in Fig.~\ref{fig:reconstruction_contrast_1}, our method reconstructs only the image regions directly associated with the target concepts. In contrast, the CEM baseline also reconstructs irrelevant background content (e.g., the text behind the person), indicating that its embeddings entangle unintended features. 
We further evaluate the localization ability of the concept tokens derived from our proposed Concept-SAE model quantitatively. For the CelebA dataset, we construct two binary masks that separately cover the face region and the background. We then compute the reconstruction error of the vision model features within each region. To quantitatively measure localization, we define the \textbf{Localization Ratio} as the ratio between the MSE of the background and that of the face:
\begin{equation}
    \text{LocR} = \dfrac{\text{MSE}\left((h - \hat{h}_{concept}) \odot M_{background}\right)}{\text{MSE}\left((h - \hat{h}_{concept}) \odot M_{face}\right)}.
\end{equation}
where \(M_{background}\) and \(M_{face}\) are the binary masks for background and face part respectively. Since only facial concepts are used to train the SAE, a higher Localization Ratio shows the facial part is reconstructed better than the background part, which indicates a stronger ability to disentangle the interested facial concepts from the background and preserve the spatial localization of the facial concepts. We compare our Concept-SAE with the CEM on shallow layers of ResNet, where localization information is still preserved. Deeper layers are excluded from evaluation, as their features are spatially fused and lack clear localization. As shown in Table~\ref{tab:localization}, our Concept-SAE consistently achieves higher localization ratios than CEM, demonstrating superior concept localization.
This highlights our method’s superior ability to isolate and faithfully represent localized, semantically meaningful concepts. Moreover, introducing supervised concept tokens does not diminish the capacity of free tokens. As shown in Fig.~\ref{fig:free_token_part}, we can still get free tokens with rich semantic information through manual search. Therefore, in our proposed Concept-SAE framework, the free tokenizer and aggregator continue to capture residual, unconstrained features, enabling semantic analysis beyond predefined concepts.

\begin{table}[!t]
\centering
\caption{Localization ratio of our proposed \textbf{Concept-SAE} and CEM. A higher localization ratio indicates the concept derived has better localization characteristics and is better disentangled.}
\label{tab:localization}
\resizebox{0.6\linewidth}{!}{
\begin{tabular}{c|cc}
\toprule
\textbf{Model layer} & \textbf{LocR (Ours)} & \textbf{LocR (CEM)}\\ %
\midrule
ResNet-18 layer 5 & 1.472 & 1.019 \\
ResNet-18 layer 7 & 1.402 & 0.982 \\
ResNet-18 layer 9 & 1.395 & 1.002 \\
\bottomrule
\end{tabular}}
\end{table}

\begin{table*}[!t]
\centering
\caption{Information entropy of the concept score for different layers of the vision models. Higher information entropy indicates the output feature of that layer is more ambiguous. {\color{red}Red} and {\color{blue}blue} numbers show the {\color{red}increase} and {\color{blue}decrease} in entropy compared to the average of the original samples. The standard deviations of the calculated concept score entropy for each layer is below 0.001.}
\label{tab:entropy}
\resizebox{0.9\linewidth}{!}{
\begin{tabular}{c|c|cccc}
\toprule
\multirow{2}{*}{\textbf{Dataset}} & \multirow{2}{*}{\textbf{Model layer}} & \multicolumn{4}{c}{\textbf{Concept score entropy}} \\ 
\cmidrule(lr){3-6}
    & & All pred. & Correct pred. & Incorrect pred. & Adversarial pred. \\ 
\midrule
\multirow{13}{*}{CelebA} & ViT layer 0 & 0.249 & 0.248 & 0.273\hspace{0.5em}({\color{red}+ 0.024\hspace{0.2em}\(\uparrow\)}) &  0.320\hspace{0.5em}({\color{red}+ 0.071\hspace{0.2em}\(\uparrow\)}) \\
&ViT layer 2 & 0.234 & 0.234 & 0.257\hspace{0.5em}({\color{red}+ 0.023\hspace{0.2em}\(\uparrow\)}) & 0.353\hspace{0.5em}({\color{red}+ 0.119\hspace{0.2em}\(\uparrow\)}) \\
&ViT layer 4 & 0.223 & 0.222 & 0.246\hspace{0.5em}({\color{red}+ 0.023\hspace{0.2em}\(\uparrow\)}) & 0.379\hspace{0.5em}({\color{red}+ 0.156\hspace{0.2em}\(\uparrow\)}) \\
&ViT layer 6 & 0.216 & 0.216 & 0.235\hspace{0.5em}({\color{red}+ 0.019\hspace{0.2em}\(\uparrow\)}) & 0.346\hspace{0.5em}({\color{red}+ 0.130\hspace{0.2em}\(\uparrow\)}) \\
&ViT layer 8 & 0.197 & 0.197 & 0.212\hspace{0.5em}({\color{red}+ 0.015\hspace{0.2em}\(\uparrow\)}) & 0.291\hspace{0.5em}({\color{red}+ 0.094\hspace{0.2em}\(\uparrow\)}) \\
&ViT layer 10 & 0.197 & 0.196 & 0.203\hspace{0.5em}({\color{red}+ 0.007\hspace{0.2em}\(\uparrow\)}) & 0.256\hspace{0.5em}({\color{red}+ 0.059\hspace{0.2em}\(\uparrow\)}) \\
&ViT layer 11 & 0.208 & 0.207 & 0.221\hspace{0.5em}({\color{red}+ 0.013\hspace{0.2em}\(\uparrow\)}) & 0.284\hspace{0.5em}({\color{red}+ 0.076\hspace{0.2em}\(\uparrow\)}) \\
\cmidrule(lr){2-6} 
&ResNet-18 layer 5 & 0.289 & 0.287 & 0.320\hspace{0.5em}({\color{red}+ 0.031\hspace{0.2em}\(\uparrow\)}) & 0.298\hspace{0.5em}({\color{red}+ 0.009\hspace{0.2em}\(\uparrow\)}) \\
&ResNet-18 layer 7 & 0.284 & 0.281 & 0.313\hspace{0.5em}({\color{red}+ 0.029\hspace{0.2em}\(\uparrow\)}) & 0.294\hspace{0.5em}({\color{red}+ 0.010\hspace{0.2em}\(\uparrow\)}) \\
&ResNet-18 layer 9 & 0.279 & 0.277 & 0.307\hspace{0.5em}({\color{red}+ 0.028\hspace{0.2em}\(\uparrow\)}) & 0.315\hspace{0.5em}({\color{red}+ 0.036\hspace{0.2em}\(\uparrow\)}) \\
&ResNet-18 layer 12 & 0.277 & 0.274 & 0.304\hspace{0.5em}({\color{red}+ 0.027\hspace{0.2em}\(\uparrow\)}) & 0.305\hspace{0.5em}({\color{red}+ 0.028\hspace{0.2em}\(\uparrow\)}) \\
&ResNet-18 layer 14 & 0.274 & 0.271 & 0.302\hspace{0.5em}({\color{red}+ 0.028\hspace{0.2em}\(\uparrow\)}) & 0.324\hspace{0.5em}({\color{red}+ 0.050\hspace{0.2em}\(\uparrow\)}) \\
&ResNet-18 layer 17 & 0.253 & 0.249 & 0.289\hspace{0.5em}({\color{red}+ 0.036\hspace{0.2em}\(\uparrow\)}) & 0.304\hspace{0.5em}({\color{red}+ 0.051\hspace{0.2em}\(\uparrow\)}) \\
\midrule
\multirow{13}{*}{Imagenet} & ViT layer 0 & 0.225 & 0.225 & 0.217\hspace{0.5em}({\color{blue}- 0.008\hspace{0.2em}\(\downarrow\)}) &  0.171\hspace{0.5em}({\color{blue}- 0.054\hspace{0.2em}\(\downarrow\)}) \\
&ViT layer 2 & 0.210 & 0.210 & 0.213\hspace{0.5em}({\color{red}+ 0.003\hspace{0.2em}\(\uparrow\)}) & 0.170\hspace{0.5em}({\color{blue}- 0.040\hspace{0.2em}\(\downarrow\)}) \\
&ViT layer 4 & 0.198 & 0.198 & 0.207\hspace{0.5em}({\color{red}+ 0.009\hspace{0.2em}\(\uparrow\)}) & 0.213\hspace{0.5em}({\color{red}+ 0.015\hspace{0.2em}\(\uparrow\)}) \\
&ViT layer 6 & 0.180 & 0.180 & 0.206\hspace{0.5em}({\color{red}+ 0.026\hspace{0.2em}\(\uparrow\)}) & 0.187\hspace{0.5em}({\color{red}+ 0.007\hspace{0.2em}\(\uparrow\)}) \\
&ViT layer 8 & 0.163 & 0.163 & 0.184\hspace{0.5em}({\color{red}+ 0.021\hspace{0.2em}\(\uparrow\)}) & 0.174\hspace{0.5em}({\color{red}+ 0.011\hspace{0.2em}\(\uparrow\)}) \\
&ViT layer 10 & 0.158 & 0.158 & 0.160\hspace{0.5em}({\color{red}+ 0.002\hspace{0.2em}\(\uparrow\)}) & 0.165\hspace{0.5em}({\color{red}+ 0.007\hspace{0.2em}\(\uparrow\)}) \\
&ViT layer 11 & 0.166 & 0.166 & 0.186\hspace{0.5em}({\color{red}+ 0.020\hspace{0.2em}\(\uparrow\)}) & 0.192\hspace{0.5em}({\color{red}+ 0.026\hspace{0.2em}\(\uparrow\)}) \\
\cmidrule(lr){2-6} 
\cmidrule(lr){2-6}
&ResNet-18 layer 5 & 0.224 & 0.224 & 0.226\hspace{0.5em}({\color{red}+ 0.002\hspace{0.2em}\(\uparrow\)}) & 0.235\hspace{0.5em}({\color{red}+ 0.011\hspace{0.2em}\(\uparrow\)}) \\
&ResNet-18 layer 7 & 0.215 & 0.214 & 0.219\hspace{0.5em}({\color{red}+ 0.004\hspace{0.2em}\(\uparrow\)}) & 0.227\hspace{0.5em}({\color{red}+ 0.012\hspace{0.2em}\(\uparrow\)}) \\
&ResNet-18 layer 9 & 0.185 & 0.185 & 0.191\hspace{0.5em}({\color{red}+ 0.006\hspace{0.2em}\(\uparrow\)}) & 0.186\hspace{0.5em}({\color{red}+ 0.001\hspace{0.2em}\(\uparrow\)}) \\
&ResNet-18 layer 12 & 0.190 & 0.188 & 0.196\hspace{0.5em}({\color{red}+ 0.006\hspace{0.2em}\(\uparrow\)}) & 0.192\hspace{0.5em}({\color{red}+ 0.002\hspace{0.2em}\(\uparrow\)}) \\
&ResNet-18 layer 14 & 0.188 & 0.188 & 0.195\hspace{0.5em}({\color{red}+ 0.007\hspace{0.2em}\(\uparrow\)}) & 0.203\hspace{0.5em}({\color{red}+ 0.015\hspace{0.2em}\(\uparrow\)}) \\
&ResNet-18 layer 17 & 0.169 & 0.169 & 0.175\hspace{0.5em}({\color{red}+ 0.006\hspace{0.2em}\(\uparrow\)}) & 0.182\hspace{0.5em}({\color{red}+ 0.013\hspace{0.2em}\(\uparrow\)}) \\
\bottomrule
\end{tabular}}
\end{table*}
\begin{table*}[!t]
\centering
\caption{Comparison between methods in terms of adversarial sample detection AUC against FGSM using ImageNet and CelebA datasets.}
\label{tab:adv_detection}
\resizebox{0.8\linewidth}{!}{
\begin{tabular}{c|c|cccccc|c}
\toprule
\multirow{2}{*}{\textbf{Dataset}} & \multirow{2}{*}{\textbf{Model}} & \multicolumn{7}{c}{\textbf{AUC score of adversarial sample detection}} \\ 
\cmidrule(lr){3-9}
    & & JPEG & Random & Deflect & Denoise & VLAD & FS & Ours \\ 
\midrule
\multirow{2}{*}{CelebA} & ViT & 0.832 & 0.807 & 0.501 & 0.512 & 0.901  & 0.622 & \colorbox{gray!24}{\textbf{0.974}} \\
&ResNet-18 & 0.755 & 0.767 & 0.493 & 0.515  & 0.896 & 0.570 & \colorbox{gray!24}{\textbf{0.940}}\\
\midrule
\multirow{2}{*}{ImageNet} & ViT & 0.693 & 0.604 & 0.493 & 0.532 & 0.876 & 0.475 & \colorbox{gray!24}{\textbf{0.948}}\\
&ResNet-18 & 0.733 & 0.660 & 0.493 & 0.672 & 0.828 & 0.533 & \colorbox{gray!24}{\textbf{0.869}}\\
\bottomrule
\end{tabular}}
\end{table*}

\subsection{RQ2: Diagnostic Utility via Detection}

For RQ2, we test whether the fidelity of the Concept-SAE representations (concept score) is high enough to enable the explanation of prediction failures, especially adversarial samples. A representation is useful for this if its state faithfully reflects the model's internal processing.
We examine the different patterns of concept scores for correctly predicted and mispredicted images. We also add adversarial images using FGSM~\cite{goodfellow2014explaining} to the experiment. We interpret each concept score as the probability that the specific concept is represented internally by the model. To quantify reliability, we compute the entropy of concept scores across image features from different vision model layers. Higher entropy indicates that the extracted feature of the vision model is ambiguous and less reliable, while lower entropy reflects more confident concept usage of the vision model. As shown in Table~\ref{tab:entropy}, for each layer of the vision model, incorrect predictions and predictions for adversarial samples consistently exhibit higher entropy in concept scores, suggesting that uncertain concept activations are a key factor of model failures.

To further evaluate the robustness of the approach, we use a simple threshold-based binary classifier on the entropy of the concept score to decide whether an image is an adversarial sample. For our method, we use the entropy of the concept score from ViT layer 4 and ResNet-18 layer 9. The classification performance is quantified using the Area Under the Curve (AUC) metric. We benchmarked our detection method against several established baselines, including JPEG~\cite{das2018JPEG}, Randomization~\cite{xie2017Randomization}, Deflection~\cite{prakash2018deflect}, Denoise~\cite{salman2020denoised}, VLAD~\cite{mumcu2024VLAD} and Feature Squeezing~\cite{xu2017FS}. As detailed in Table~\ref{tab:adv_detection}, our proposed method consistently demonstrates superior detection capabilities of adversarial images, outperforming all previously established baseline techniques. Therefore proving our proposed concept score faithfully reflects the model's internal processing.

\subsection{RQ3: Diagnostic Utility via Controllability}

\begin{figure*}[!t]
    \centering
    \includegraphics[width=\linewidth]{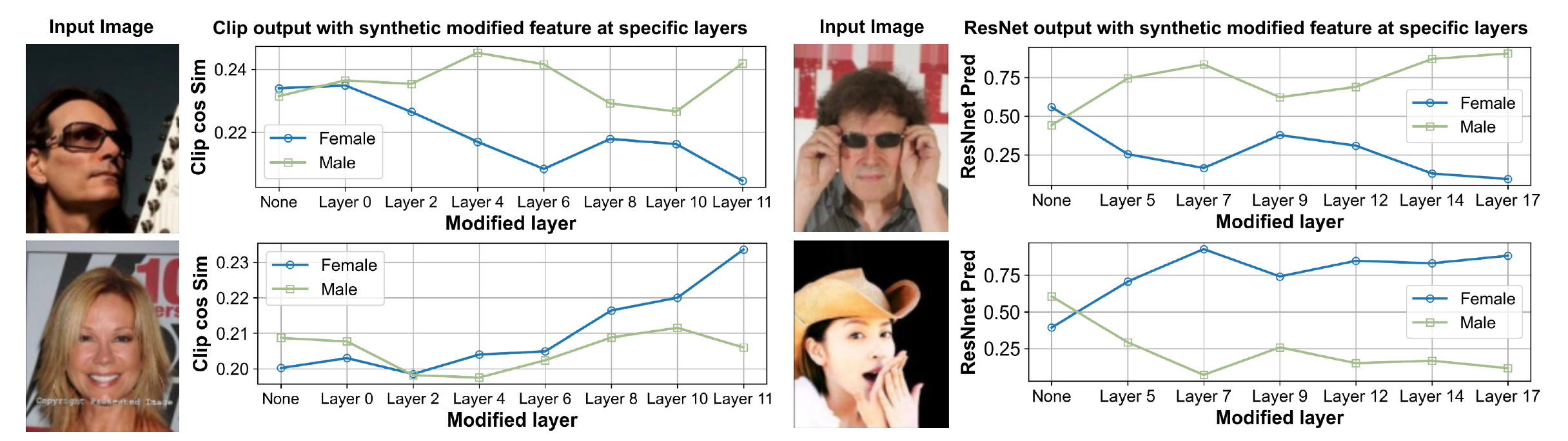}
    \caption{We modify the concept score and generate synthetic features through aggregator at different layers of the vision model. For male figures, we increase the concept score of \textit{beard}, \textit{adam's apple} to \(1.0\). For female figures, we decrease the concept score of \textit{beard}, \textit{adam's apple} to \(0.0\). Then we examine how the vision models perform with these new features on failure images.}
    \label{fig:modify_score_example}
\vspace{-0.2cm}
\end{figure*}

For RQ3, we test whether the fidelity of the Concept-SAE representation is high enough to enable reliable, controllable counterfactual editing on vision model output. We intervene directly on the concept scores to modify the final output of the vision model. Specifically, we adjust selected concept scores to form a modified concept score vector, for instance, set the concept score of \textit{beard} to zero. Then, the modified concept scores are passed through the aggregator with the original concept tokens to generate a counterfactual feature representation. We further substitute the original vision model feature with this generated counterfactual feature as the input of the corresponding layer in the vision model so that we can modify the final output of the vision model. On CelebA misclassifications, for example, \textit{male} images incorrectly predicted as \textit{female} often exhibit insufficient activations of male-associated concepts such as beard, mustache, and adam's apple. By increasing the scores of these concepts and feed the synthetic representation back to the vision model, the prediction is corrected to male; conversely, reducing them corrects the opposite errors. As shown in Fig.~\ref{fig:modify_score_example}, we find that interventions on deeper ViT layers are more effective, while in ResNet both shallow and deep layers yield strong corrections.
These results demonstrate that Concept-SAE not only identifies ambiguous concept activations as the cause of errors but also enables direct, causal interventions to repair predictions. The capability is not possible in prior SAE methods without explicit concept scores.

\begin{table*}[!t]
\centering
\caption{Accuracy of the model on \textbf{adversarial samples} after finetuning. JS distance indicates the difference between the concept scores of the original samples and those of the adversarial samples in each layer. {\color{red}Red} and {\color{blue}Blue} numbers show the increase in accuracy and decrease in concept score JS distance for adversarial samples after adversarial finetuning. \colorbox{yellow!25}{Yellow background} shows the top-3 values in that column.}
\label{tab:adv_improve}
\resizebox{0.9\linewidth}{!}{
\begin{tabular}{c|c|lcc}
\toprule
\textbf{Dataset} & \makecell{\textbf{Finetuned} \\ \textbf{Model Layer}} & \makecell{\textbf{Adv Sample} \\ \textbf{Accuracy}} &  \makecell{\textbf{JS distance} \\ \textbf{(before finetune)}} & \makecell{\textbf{JS distance} \\ \textbf{(after finetune)}} \\
\midrule
\multirow{17}{*}{CelebA} & None (ViT) & \hspace{3.4em}70.05 \% & - & - \\
\cmidrule(lr){2-5}
&ViT layer 0 & \hspace{0.8em}\cellcolor{yellow!25}87.67\%\hspace{0.5em}({\color{red}+ 17.62\%\hspace{0.2em}\(\uparrow\)}) & \cellcolor{yellow!25}0.178 & 0.172\hspace{0.5em}({\color{blue}- 0.006\hspace{0.2em}\(\downarrow\)}) \\
&ViT layer 2 & \hspace{0.8em}80.03\%\hspace{0.5em}({\color{red}+ 9.98\%\hspace{0.2em}\(\uparrow\)})  & 0.137 & 0.103\hspace{0.5em}({\color{blue}- 0.034\hspace{0.2em}\(\downarrow\)}) \\
&ViT layer 4 & \hspace{0.8em}\cellcolor{yellow!25}87.03\%\hspace{0.5em}({\color{red}+ 16.98\%\hspace{0.2em}\(\uparrow\)}) & \cellcolor{yellow!25}0.225 & 0.173\hspace{0.5em}({\color{blue}- 0.052\hspace{0.2em}\(\downarrow\)}) \\
&ViT layer 6 & \hspace{0.8em}\cellcolor{yellow!25}83.06\%\hspace{0.5em}({\color{red}+ 13.01\%\hspace{0.2em}\(\uparrow\)}) & \cellcolor{yellow!25}0.160 & 0.125\hspace{0.5em}({\color{blue}- 0.035\hspace{0.2em}\(\downarrow\)}) \\
&ViT layer 8 & \hspace{0.8em}80.36\%\hspace{0.5em}({\color{red}+ 10.31\%\hspace{0.2em}\(\uparrow\)}) & 0.129 & 0.100\hspace{0.5em}({\color{blue}- 0.029\hspace{0.2em}\(\downarrow\)}) \\
&ViT layer 10 & \hspace{0.8em}76.88\%\hspace{0.5em}({\color{red}+ 6.83\%\hspace{0.2em}\(\uparrow\)}) & 0.121 & 0.064\hspace{0.5em}({\color{blue}- 0.057\hspace{0.2em}\(\downarrow\)}) \\
&ViT layer 11 & \hspace{0.8em}74.33\%\hspace{0.5em}({\color{red}+ 4.28\%\hspace{0.2em}\(\uparrow\)}) & 0.119 & 0.062\hspace{0.5em}({\color{blue}- 0.057\hspace{0.2em}\(\downarrow\)}) \\
\cmidrule(lr){2-5}
&All layers (ViT) & \hspace{0.8em}93.32\%\hspace{0.5em}({\color{red}+ 23.27\%\hspace{0.2em}\(\uparrow\)}) & - & - \\
\cmidrule(lr){2-5}
\morecmidrules
\cmidrule(lr){2-5}
&None (ResNet) & \hspace{3.4em}39.75\% & - & - \\
\cmidrule(lr){2-5}
&ResNet-18 layer 5 & \hspace{0.8em}55.45\%\hspace{0.5em}({\color{red}+ 15.70\%\hspace{0.2em}\(\uparrow\)}) & 0.066 & 0.035\hspace{0.5em}({\color{blue}- 0.031\hspace{0.2em}\(\downarrow\)}) \\
&ResNet-18 layer 7 & \hspace{0.8em}61.19\%\hspace{0.5em}({\color{red}+ 21.44\%\hspace{0.2em}\(\uparrow\)}) & 0.080 & 0.049\hspace{0.5em}({\color{blue}- 0.031\hspace{0.2em}\(\downarrow\)}) \\
&ResNet-18 layer 9 & \hspace{0.8em}60.51\%\hspace{0.5em}({\color{red}+ 20.76\%\hspace{0.2em}\(\uparrow\)}) & 0.087 & 0.054\hspace{0.5em}({\color{blue}- 0.033\hspace{0.2em}\(\downarrow\)}) \\
&ResNet-18 layer 12 & \hspace{0.8em}\cellcolor{yellow!25}63.53\%\hspace{0.5em}({\color{red}+ 23.78\%\hspace{0.2em}\(\uparrow\)}) & \cellcolor{yellow!25}0.095 & 0.072\hspace{0.5em}({\color{blue}- 0.023\hspace{0.2em}\(\downarrow\)}) \\
&ResNet-18 layer 14 & \hspace{0.8em}\cellcolor{yellow!25}67.09\%\hspace{0.5em}({\color{red}+ 27.34\%\hspace{0.2em}\(\uparrow\)}) & \cellcolor{yellow!25}0.098 & 0.087\hspace{0.5em}({\color{blue}- 0.012\hspace{0.2em}\(\downarrow\)}) \\
&ResNet-18 layer 17 & \hspace{0.8em}\cellcolor{yellow!25}68.08\%\hspace{0.5em}({\color{red}+ 28.33\%\hspace{0.2em}\(\uparrow\)}) & \cellcolor{yellow!25}0.109 & 0.106\hspace{0.5em}({\color{blue}- 0.003\hspace{0.2em}\(\downarrow\)}) \\
\cmidrule(lr){2-5}
&All layers (ResNet) & \hspace{0.8em}80.52\%\hspace{0.5em}({\color{red}+ 40.77\%\hspace{0.2em}\(\uparrow\)}) & - & - \\
\midrule
\multirow{17}{*}{Imagenet} & None (ViT) & \hspace{3.4em}11.78 \% & - & - \\
\cmidrule(lr){2-5}
&ViT layer 0 & \hspace{0.8em}\cellcolor{yellow!25}28.24\%\hspace{0.5em}({\color{red}+ 16.46\%\hspace{0.2em}\(\uparrow\)}) & \cellcolor{yellow!25}0.070 & 0.052\hspace{0.5em}({\color{blue}- 0.018\hspace{0.2em}\(\downarrow\)}) \\
&ViT layer 2 & \cellcolor{yellow!25}\hspace{0.8em}29.23\%\hspace{0.5em}({\color{red}+ 17.45\%\hspace{0.2em}\(\uparrow\)})  & \cellcolor{yellow!25}0.065 & 0.054\hspace{0.5em}({\color{blue}- 0.011\hspace{0.2em}\(\downarrow\)}) \\
&ViT layer 4 & \hspace{0.8em}\cellcolor{yellow!25}29.78\%\hspace{0.5em}({\color{red}+ 18.00\%\hspace{0.2em}\(\uparrow\)}) & \cellcolor{yellow!25}0.067 & 0.043\hspace{0.5em}({\color{blue}- 0.024\hspace{0.2em}\(\downarrow\)}) \\
&ViT layer 6 & \hspace{0.8em}24.81\%\hspace{0.5em}({\color{red}+ 13.03\%\hspace{0.2em}\(\uparrow\)}) & 0.052 & 0.040\hspace{0.5em}({\color{blue}- 0.012\hspace{0.2em}\(\downarrow\)}) \\
&ViT layer 8 & \hspace{0.8em}25.77\%\hspace{0.5em}({\color{red}+ 13.99\%\hspace{0.2em}\(\uparrow\)}) & 0.051 & 0.041\hspace{0.5em}({\color{blue}- 0.010\hspace{0.2em}\(\downarrow\)}) \\
&ViT layer 10 & \hspace{0.8em}27.84\%\hspace{0.5em}({\color{red}+ 16.06\%\hspace{0.2em}\(\uparrow\)}) & 0.053 & 0.045\hspace{0.5em}({\color{blue}- 0.008\hspace{0.2em}\(\downarrow\)}) \\
&ViT layer 11 & \hspace{0.8em}25.24\%\hspace{0.5em}({\color{red}+ 13.46\%\hspace{0.2em}\(\uparrow\)}) & 0.048 & 0.040\hspace{0.5em}({\color{blue}- 0.008\hspace{0.2em}\(\downarrow\)}) \\
\cmidrule(lr){2-5}
&All layers (ViT) & \hspace{0.8em}34.98\%\hspace{0.5em}({\color{red}+ 23.20\%\hspace{0.2em}\(\uparrow\)}) & - & - \\
\cmidrule(lr){2-5}
\morecmidrules
\cmidrule(lr){2-5}
&None (ResNet) & \hspace{3.4em}9.51\% & - & - \\
\cmidrule(lr){2-5}
&ResNet-18 layer 5 & \hspace{0.8em}13.87\%\hspace{0.5em}({\color{red}+ 4.36\%\hspace{0.2em}\(\uparrow\)}) & 0.030 & 0.017\hspace{0.5em}({\color{blue}- 0.013\hspace{0.2em}\(\downarrow\)}) \\
&ResNet-18 layer 7 & \hspace{0.8em}15.67\%\hspace{0.5em}({\color{red}+ 6.16\%\hspace{0.2em}\(\uparrow\)}) & 0.034 & 0.022\hspace{0.5em}({\color{blue}- 0.012\hspace{0.2em}\(\downarrow\)}) \\
&ResNet-18 layer 9 & \hspace{0.8em}\cellcolor{yellow!25}17.20\%\hspace{0.5em}({\color{red}+ 7.69\%\hspace{0.2em}\(\uparrow\)}) & \cellcolor{yellow!25}0.039 & 0.027\hspace{0.5em}({\color{blue}- 0.012\hspace{0.2em}\(\downarrow\)}) \\
&ResNet-18 layer 12 & \hspace{0.8em}\cellcolor{yellow!25}16.59\%\hspace{0.5em}({\color{red}+ 7.08\%\hspace{0.2em}\(\uparrow\)}) & 0.036 & 0.026\hspace{0.5em}({\color{blue}- 0.010\hspace{0.2em}\(\downarrow\)}) \\
&ResNet-18 layer 14 & \hspace{0.8em}14.89\%\hspace{0.5em}({\color{red}+ 5.38\%\hspace{0.2em}\(\uparrow\)}) & \cellcolor{yellow!25}0.038 & 0.027\hspace{0.5em}({\color{blue}- 0.011\hspace{0.2em}\(\downarrow\)}) \\
&ResNet-18 layer 17 & \hspace{0.8em}\cellcolor{yellow!25}17.33\%\hspace{0.5em}({\color{red}+ 7.82\%\hspace{0.2em}\(\uparrow\)}) & \cellcolor{yellow!25}0.047 & 0.041\hspace{0.5em}({\color{blue}- 0.006\hspace{0.2em}\(\downarrow\)}) \\
\cmidrule(lr){2-5}
&All layers (ResNet) & \hspace{0.8em}34.71\%\hspace{0.5em}({\color{red}+ 25.20\%\hspace{0.2em}\(\uparrow\)}) & - & - \\
\bottomrule
\end{tabular}}
\vspace{-0.2cm}
\end{table*}

\subsection{RQ4: Diagnostic Utility via Stability Diagnosis}

For RQ4, we test if the semantic alignment of our concept tokens is faithful enough to serve as a semantic stability test. The goal is not to propose a new adversarial defense, but to validate whether our representations can be used to meaningfully reflect the vulnerabilities of models under worst-case perturbations.
To investigate this, we generate adversarial examples using FGSM~\cite{goodfellow2014explaining} and monitor changes in concept scores across the model hierarchy. An entropy-based analysis reveals two complementary effects: (1) some layers become excessively confident in a small set of concepts, as evidenced by reduced entropy, suggesting that the model latches onto spurious signals amplified by the perturbation; (2) other layers display the opposite behavior, with increased entropy reflecting heightened uncertainty and confusion in concept extraction. These opposing tendencies, confidence collapse and semantic diffusion, jointly illustrate how adversarial perturbations erode the consistency of internal representations and disrupt the alignment between concepts and predictions.
To more precisely quantify such changes, we compute the Jensen–Shannon (JS) distance~\cite{lin2002divergence} between the concept score distributions of clean and adversarial samples. Larger JS distances indicate stronger distributional shifts encountering adversarial samples and thus higher vulnerability. Based on this observation, we hypothesize that layers with higher JS distances are most fragile under attack.
We validate this hypothesis through layer-wise finetuning. For each layer, we freeze the rest layers and retrain the chosen layer using mixed clean and adversarial samples for two epochs. As shown in Table~\ref{tab:adv_improve}, layers identified with larger JS distances consistently yield greater robustness improvements after finetuning compared to less vulnerable layers. The success of this experiment validates that our concept representations provide a high-fidelity semantic stability test. They faithfully report how and where adversarial attacks distort the model's internal conceptual understanding, offering a precise, validated diagnostic tool for localizing vulnerabilities.

\section{Limitations}

Despite its effectiveness, \textbf{Concept-SAE} has several limitations. First, the accuracy of interpretation depends on the quality of the generated concept supervision. Noise or inaccuracies from the vision-language model or the segmentation model may lead to imprecise concept identification and risk imposing misaligned interpretations onto the target model, though this issue is expected to diminish as foundation models continue to improve. Second, the reliance on spatial masks makes the framework particularly suited for concepts corresponding to localizable objects. Extending this approach to more abstract, textural, or globally distributed concepts where spatial segmentation is inherently difficult remains an important direction for future research.

\section{Conclusion}

In this paper, we introduced Concept-SAE, a framework designed to extend Sparse Autoencoders from passive feature discovery toward active inspection of how vision models represent human-defined concepts. Our core contribution is a hybrid disentanglement strategy that separates an activation subspace into semantically aligned concept tokens, trained with dual supervision on concept existence and spatial localization, and unconstrained free tokens that preserve the exploratory capability of standard SAEs. Through extensive experiments, we demonstrated that Concept-SAE produces faithful, localized, and disentangled concept representations.
We further validated the interface through three evaluations: (1) a detection test, where the entropy of concept score is used to classify adversarial image samples, (2) a controllability test, where controlled counterfactual editing reveals how manipulating concept tokens influences model predictions, and (3) a stability test, where adversarial perturbations expose where and how concept representations become vulnerable.
Overall, Concept-SAE provides a validated and controllable interface that complements traditional SAEs, enabling researchers to move beyond passive observation and actively probe the role of specific concepts in a model’s internal mechanisms.

    \subsubsection*{Usage of Large Language Model.}
We used Large language model (LLM) to aid and polish writing. We did not use LLM for other purposes.

%
%
%
\bibliographystyle{splncs04}
\bibliography{main}

@String(ICCV= {Int. Conf. Comput. Vis.})

@String(ICCV  = {ICCV})

@inproceedings{radford2021learning,
  title={Learning transferable visual models from natural language supervision},
  author={Radford, Alec and Kim, Jong Wook and Hallacy, Chris and Ramesh, Aditya and Goh, Gabriel and Agarwal, Sandhini and Sastry, Girish and Askell, Amanda and Mishkin, Pamela and Clark, Jack and others},
  booktitle={International conference on machine learning},
  pages={8748--8763},
  year={2021},
  organization={PmLR}
}

@article{dosovitskiy2020image,
  title={An image is worth 16x16 words: Transformers for image recognition at scale},
  author={Dosovitskiy, Alexey and Beyer, Lucas and Kolesnikov, Alexander and Weissenborn, Dirk and Zhai, Xiaohua and Unterthiner, Thomas and Dehghani, Mostafa and Minderer, Matthias and Heigold, Georg and Gelly, Sylvain and others},
  journal={arXiv preprint arXiv:2010.11929},
  year={2020}
}

@inproceedings{he2016deep,
  title={Deep residual learning for image recognition},
  author={He, Kaiming and Zhang, Xiangyu and Ren, Shaoqing and Sun, Jian},
  booktitle={Proceedings of the IEEE conference on computer vision and pattern recognition},
  pages={770--778},
  year={2016}
}

@inproceedings{deng2009imagenet,
  title={Imagenet: A large-scale hierarchical image database},
  author={Deng, Jia and Dong, Wei and Socher, Richard and Li, Li-Jia and Li, Kai and Fei-Fei, Li},
  booktitle={2009 IEEE conference on computer vision and pattern recognition},
  pages={248--255},
  year={2009},
  organization={Ieee}
}

@inproceedings{liu2015faceattributes,
  title = {Deep Learning Face Attributes in the Wild},
  author = {Liu, Ziwei and Luo, Ping and Wang, Xiaogang and Tang, Xiaoou},
  booktitle = {Proceedings of International Conference on Computer Vision (ICCV)},
  month = {December},
  year = {2015} 
}

@article{goodfellow2014explaining,
  title={Explaining and harnessing adversarial examples},
  author={Goodfellow, Ian J and Shlens, Jonathon and Szegedy, Christian},
  journal={arXiv preprint arXiv:1412.6572},
  year={2014}
}

@inproceedings{huben2023sparse,
  title={Sparse autoencoders find highly interpretable features in language models},
  author={Huben, Robert and Cunningham, Hoagy and Smith, Logan Riggs and Ewart, Aidan and Sharkey, Lee},
  booktitle={The Twelfth International Conference on Learning Representations},
  year={2023}
}

@article{olah2020zoom,
  title={Zoom in: An introduction to circuits},
  author={Olah, Chris and Cammarata, Nick and Schubert, Ludwig and Goh, Gabriel and Petrov, Michael and Carter, Shan},
  journal={Distill},
  volume={5},
  number={3},
  pages={e00024--001},
  year={2020}
}

@article{zhang2018visual,
  title={Visual interpretability for deep learning: a survey},
  author={Zhang, Quan-shi and Zhu, Song-Chun},
  journal={Frontiers of Information Technology \& Electronic Engineering},
  volume={19},
  number={1},
  pages={27--39},
  year={2018},
  publisher={Springer}
}

@inproceedings{rao2024discover,
  title={Discover-then-name: Task-agnostic concept bottlenecks via automated concept discovery},
  author={Rao, Sukrut and Mahajan, Sweta and B{\"o}hle, Moritz and Schiele, Bernt},
  booktitle={European Conference on Computer Vision},
  pages={444--461},
  year={2024},
  organization={Springer}
}

@inproceedings{ramaswamy2023overlooked,
  title={Overlooked factors in concept-based explanations: Dataset choice, concept learnability, and human capability},
  author={Ramaswamy, Vikram V and Kim, Sunnie SY and Fong, Ruth and Russakovsky, Olga},
  booktitle={Proceedings of the IEEE/CVF Conference on Computer Vision and Pattern Recognition},
  pages={10932--10941},
  year={2023}
}

@article{espinosa2023learning,
  title={Learning to receive help: Intervention-aware concept embedding models},
  author={Espinosa Zarlenga, Mateo and Collins, Katie and Dvijotham, Krishnamurthy and Weller, Adrian and Shams, Zohreh and Jamnik, Mateja},
  journal={Advances in Neural Information Processing Systems},
  volume={36},
  pages={37849--37875},
  year={2023}
}

@article{espinosa2022concept,
  title={Concept embedding models: Beyond the accuracy-explainability trade-off},
  author={Espinosa Zarlenga, Mateo and Barbiero, Pietro and Ciravegna, Gabriele and Marra, Giuseppe and Giannini, Francesco and Diligenti, Michelangelo and Shams, Zohreh and Precioso, Frederic and Melacci, Stefano and Weller, Adrian and others},
  journal={Advances in neural information processing systems},
  volume={35},
  pages={21400--21413},
  year={2022}
}

@article{rajamanoharan2024improving,
  title={Improving dictionary learning with gated sparse autoencoders},
  author={Rajamanoharan, Senthooran and Conmy, Arthur and Smith, Lewis and Lieberum, Tom and Varma, Vikrant and Kram{\'a}r, J{\'a}nos and Shah, Rohin and Nanda, Neel},
  journal={arXiv preprint arXiv:2404.16014},
  year={2024}
}

@article{o2024sparse,
  title={Sparse autoencoders enable scalable and reliable circuit identification in language models},
  author={O'Neill, Charles and Bui, Thang},
  journal={arXiv preprint arXiv:2405.12522},
  year={2024}
}

@article{marks2023interpreting,
  title={Interpreting reward models in rlhf-tuned language models using sparse autoencoders},
  author={Marks, Luke and Abdullah, Amir and Mendez, Luna and Arike, Rauno and Torr, Philip and Barez, Fazl},
  year={2023}
}

@inproceedings{sharkey2022taking,
  title={Taking features out of superposition with sparse autoencoders},
  author={Sharkey, Lee and Braun, Dan and Millidge, Beren},
  booktitle={AI Alignment Forum},
  volume={6},
  pages={12--13},
  year={2022}
}

@article{gorton2024missing,
  title={The missing curve detectors of inceptionv1: Applying sparse autoencoders to inceptionv1 early vision},
  author={Gorton, Liv},
  journal={arXiv preprint arXiv:2406.03662},
  year={2024}
}

@article{kissane2024interpreting,
  title={Interpreting attention layer outputs with sparse autoencoders},
  author={Kissane, Connor and Krzyzanowski, Robert and Bloom, Joseph Isaac and Conmy, Arthur and Nanda, Neel},
  journal={arXiv preprint arXiv:2406.17759},
  year={2024}
}

@article{yeh2020completeness,
  title={On completeness-aware concept-based explanations in deep neural networks},
  author={Yeh, Chih-Kuan and Kim, Been and Arik, Sercan and Li, Chun-Liang and Pfister, Tomas and Ravikumar, Pradeep},
  journal={Advances in neural information processing systems},
  volume={33},
  pages={20554--20565},
  year={2020}
}

@article{shu2025survey,
  title={A survey on sparse autoencoders: Interpreting the internal mechanisms of large language models},
  author={Shu, Dong and Wu, Xuansheng and Zhao, Haiyan and Rai, Daking and Yao, Ziyu and Liu, Ninghao and Du, Mengnan},
  journal={arXiv preprint arXiv:2503.05613},
  year={2025}
}

@article{stevens2025sparse,
  title={Sparse autoencoders for scientifically rigorous interpretation of vision models},
  author={Stevens, Samuel and Chao, Wei-Lun and Berger-Wolf, Tanya and Su, Yu},
  journal={arXiv preprint arXiv:2502.06755},
  year={2025}
}

@article{olson2025probing,
  title={Probing the Representational Power of Sparse Autoencoders in Vision Models},
  author={Olson, Matthew Lyle and Hinck, Musashi and Ratzlaff, Neale and Li, Changbai and Howard, Phillip and Lal, Vasudev and Tseng, Shao-Yen},
  journal={arXiv preprint arXiv:2508.11277},
  year={2025}
}

@article{lim2024sparse,
  title={Sparse autoencoders reveal selective remapping of visual concepts during adaptation},
  author={Lim, Hyesu and Choi, Jinho and Choo, Jaegul and Schneider, Steffen},
  journal={arXiv preprint arXiv:2412.05276},
  year={2024}
}

@article{olshausen1997sparse,
  title={Sparse coding with an overcomplete basis set: A strategy employed by V1?},
  author={Olshausen, Bruno A and Field, David J},
  journal={Vision research},
  volume={37},
  number={23},
  pages={3311--3325},
  year={1997},
  publisher={Elsevier}
}

@article{gao2024scaling,
  title={Scaling and evaluating sparse autoencoders},
  author={Gao, Leo and la Tour, Tom Dupr{\'e} and Tillman, Henk and Goh, Gabriel and Troll, Rajan and Radford, Alec and Sutskever, Ilya and Leike, Jan and Wu, Jeffrey},
  journal={arXiv preprint arXiv:2406.04093},
  year={2024}
}

@article{paulo2025sparse,
  title={Sparse autoencoders trained on the same data learn different features},
  author={Paulo, Gon{\c{c}}alo and Belrose, Nora},
  journal={arXiv preprint arXiv:2501.16615},
  year={2025}
}

@article{marks2024enhancing,
  title={Enhancing neural network interpretability with feature-aligned sparse autoencoders},
  author={Marks, Luke and Paren, Alasdair and Krueger, David and Barez, Fazl},
  journal={arXiv preprint arXiv:2411.01220},
  year={2024}
}

@article{harle2025measuring,
  title={Measuring and Guiding Monosemanticity},
  author={H{\"a}rle, Ruben and Friedrich, Felix and Brack, Manuel and W{\"a}ldchen, Stephan and Deiseroth, Bj{\"o}rn and Schramowski, Patrick and Kersting, Kristian},
  journal={arXiv preprint arXiv:2506.19382},
  year={2025}
}

@inproceedings{koh2020concept,
  title={Concept bottleneck models},
  author={Koh, Pang Wei and Nguyen, Thao and Tang, Yew Siang and Mussmann, Stephen and Pierson, Emma and Kim, Been and Liang, Percy},
  booktitle={International conference on machine learning},
  pages={5338--5348},
  year={2020},
  organization={PMLR}
}

@inproceedings{kantamnenisparse,
  title={Are Sparse Autoencoders Useful? A Case Study in Sparse Probing},
  author={Kantamneni, Subhash and Engels, Joshua and Rajamanoharan, Senthooran and Tegmark, Max and Nanda, Neel},
  booktitle={Forty-second International Conference on Machine Learning},
  year={2025},
}

@inproceedings{mudideefficient,
  title={Efficient Dictionary Learning with Switch Sparse Autoencoders},
  author={Mudide, Anish and Engels, Joshua and Michaud, Eric J and Tegmark, Max and de Witt, Christian Schroeder},
  booktitle={The Thirteenth International Conference on Learning Representations},
  year={2025},
}

@inproceedings{minegishirethinking,
  title={Rethinking Evaluation of Sparse Autoencoders through the Representation of Polysemous Words},
  author={Minegishi, Gouki and Furuta, Hiroki and Iwasawa, Yusuke and Matsuo, Yutaka},
  booktitle={The Thirteenth International Conference on Learning Representations},
  year={2025},
}

@article{lin2002divergence,
  title={Divergence measures based on the Shannon entropy},
  author={Lin, Jianhua},
  journal={IEEE Transactions on Information theory},
  volume={37},
  number={1},
  pages={145--151},
  year={2002},
  publisher={IEEE}
}

@article{olah2018building,
  title={The building blocks of interpretability},
  author={Olah, Chris and Satyanarayan, Arvind and Johnson, Ian and Carter, Shan and Schubert, Ludwig and Ye, Katherine and Mordvintsev, Alexander},
  journal={Distill},
  volume={3},
  number={3},
  pages={e10},
  year={2018}
}

@inproceedings{das2018JPEG,
  title={Shield: Fast, practical defense and vaccination for deep learning using jpeg compression},
  author={Das, Nilaksh and Shanbhogue, Madhuri and Chen, Shang-Tse and Hohman, Fred and Li, Siwei and Chen, Li and Kounavis, Michael E and Chau, Duen Horng},
  booktitle={Proceedings of the 24th ACM SIGKDD International Conference on Knowledge Discovery \& Data Mining},
  pages={196--204},
  year={2018}
}

@article{xie2017Randomization,
  title={Mitigating adversarial effects through randomization},
  author={Xie, Cihang and Wang, Jianyu and Zhang, Zhishuai and Ren, Zhou and Yuille, Alan},
  journal={arXiv preprint arXiv:1711.01991},
  year={2017}
}

@inproceedings{prakash2018deflect,
  title={Deflecting adversarial attacks with pixel deflection},
  author={Prakash, Aaditya and Moran, Nick and Garber, Solomon and DiLillo, Antonella and Storer, James},
  booktitle={Proceedings of the IEEE conference on computer vision and pattern recognition},
  pages={8571--8580},
  year={2018}
}

@inproceedings{mumcu2024VLAD,
  title={Multimodal attack detection for action recognition models},
  author={Mumcu, Furkan and Yilmaz, Yasin},
  booktitle={Proceedings of the IEEE/CVF Conference on Computer Vision and Pattern Recognition},
  pages={2967--2976},
  year={2024}
}

@article{xu2017FS,
  title={Feature squeezing: Detecting adversarial examples in deep neural networks},
  author={Xu, Weilin and Evans, David and Qi, Yanjun},
  journal={arXiv preprint arXiv:1704.01155},
  year={2017}
}

@article{salman2020denoised,
  title={Denoised smoothing: A provable defense for pretrained classifiers},
  author={Salman, Hadi and Sun, Mingjie and Yang, Greg and Kapoor, Ashish and Kolter, J Zico},
  journal={Advances in Neural Information Processing Systems},
  volume={33},
  pages={21945--21957},
  year={2020}
}
%





\end{document}